\documentclass[11pt,a4paper]{article}
\usepackage[hyperref]{acl2021}
\usepackage{nert}
\usepackage{times}
\usepackage{latexsym}

\usepackage{microtype}
\usepackage{subcaption}
\usepackage{graphicx}
\graphicspath{{images/}}
\usepackage{booktabs}
\usepackage{caption}
\usepackage{fnpct}
\usepackage{amsfonts}
\aclfinalcopy 
\def\aclpaperid{3838} 

\title{Happy Dance, Slow Clap:\\
Using Reaction GIFs to Predict Induced Affect on Twitter}
\author{Boaz Shmueli$^{1,2,3}$~,~ Soumya Ray$^2$, \and Lun-Wei Ku$^3$\\
$^1$Social Networks and Human-Centered Computing, TIGP, Academia Sinica\\
$^2$Institute of Service Science, National Tsing Hua University\\
$^3$Institute of Information Science, Academia Sinica\\
{\small{shmueli@iis.sinica.edu.tw~~soumya.ray@iss.nthu.edu.tw~~lwku@iis.sinica.edu.tw}}\\
}

\begin{document}

\maketitle
\begin{abstract}
Datasets with \textit{induced emotion} labels are scarce but of utmost importance for many NLP tasks. We present a new, automated method for collecting texts along with their \textit{induced reaction} labels. The method exploits the online use of reaction GIFs, which capture complex affective states. 
We show how to augment the data with \textit{induced emotions} and \textit{induced sentiment} labels. We use our method to create and publish ReactionGIF, a first-of-its-kind affective dataset of 30K tweets. We provide  baselines for three new tasks, including induced sentiment prediction and multilabel classification of induced emotions. Our method and dataset open new research opportunities in emotion detection and 
affective computing.
\end{abstract}
\section{Introduction}
Affective states such as emotions are an elemental part of the human condition. The automatic detection of these states
is thus an important task in affective computing, with  applications in  diverse fields including  psychology, political science, and marketing \cite{seyeditabari2018emotion}. Training machine learning algorithms for such applications  requires large yet task-specific emotion-labeled datasets \cite{bostan-klinger-2018-analysis}.

Borrowing from music \cite{gabrielsson2001emotion} and film \cite{tian2017recognizing}, one can distinguish between two reader perspectives when labeling emotions in text: \textit{perceived} emotions, which are the emotions that the reader recognizes in the  text, and  \textit{induced} emotions, which are the emotions aroused in the reader.
However, with the exception of \citet{buechel-hahn-2017-readers}, this distinction is mostly missing from the NLP literature, which focuses on the distinction between author and  reader perspectives \cite{calvo2013emotions}.

The collection of perceived emotions data is considerably simpler than induced emotions data, and presently most human-annotated emotion datasets
are labeled with perceived emotions
\citep[e.\,g.,][]{strapparava2008learning,preoctiuc2016modelling,hsu-ku-2018-socialnlp,demszky-etal-2020-goemotions}. Induced emotions data can be collected using physiological measurements or self-reporting, but both methods are complex, expensive, unreliable and cannot scale easily. 
Still, having well-classified induced emotions data is of utmost importance to dialogue systems and other applications that aim to detect, predict, or elicit a particular emotional response in  users. 
\newcite{pool-nissim-2016-distant} used distant supervision to detect induced emotions from Facebook posts by looking at the six available emoji reactions. Although this  method is automatic, it is limited both in emotional range, since the set of reactions is small and rigid,
and accuracy, because emojis are often misunderstood due to their visual ambiguity \cite{tigwell-emoji}.
\begin{figure*}[!t]
\centering
  \begin{tabular}[b]{ccc}
      \begin{subfigure}[b]{0.305\textwidth}
        \includegraphics[width=\linewidth]{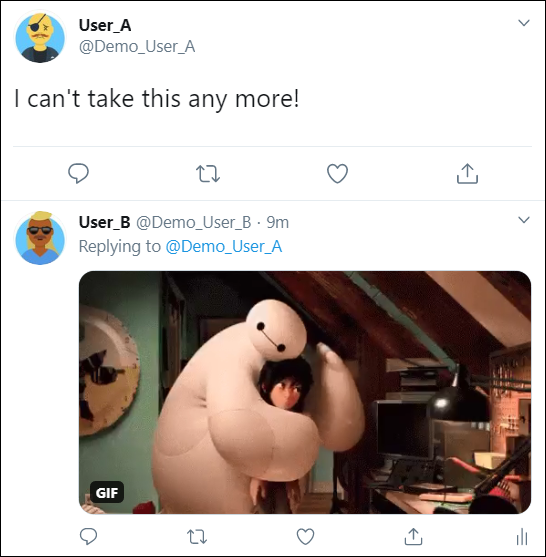}
        \caption{A root tweet (``I can't take...'') with a \textit{hug} reaction GIF reply.}
        \label{fig:2tuple}
      \end{subfigure}
    &
      \begin{subfigure}[b]{0.29\textwidth}
        \includegraphics[width=\linewidth]{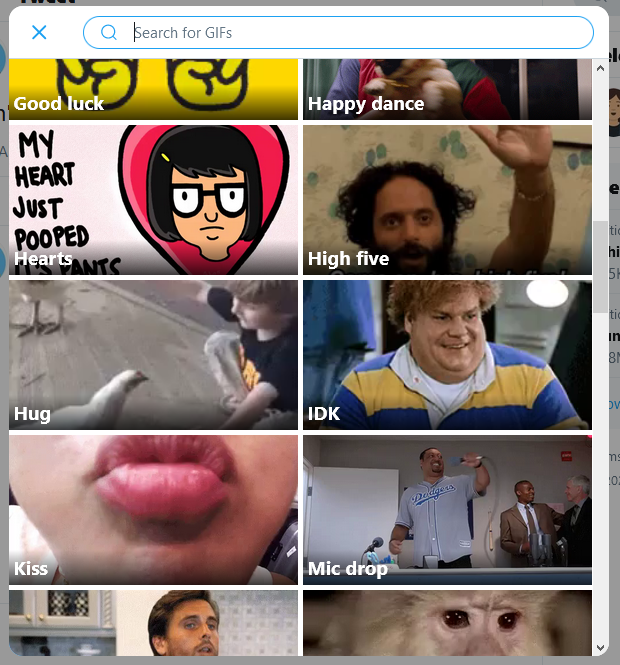}
        \caption{The reaction categories menu offers 43 categories (\textit{hug}, \textit{kiss}, ...).}
        \label{fig:categories}
      \end{subfigure}
    &
      \begin{subfigure}[b]{0.29\textwidth}
        \includegraphics[width=\linewidth]{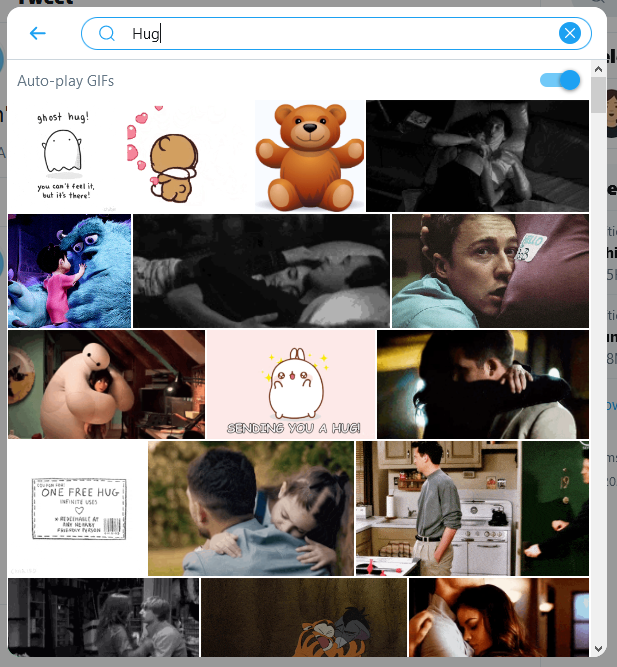}
        \caption{The top reaction GIFs offered to the user from the \textit{hug} category.}
        \label{fig:hugs}
      \end{subfigure}
  \end{tabular}
  \label{fig:ABC}
  \caption{How reaction GIFs are used (left) and inserted (middle, right) on Twitter.}
\end{figure*}

To overcome these drawbacks, we propose a new method that innovatively exploits the use of reaction GIFs in online conversations.
Reaction GIFs are effective because they 
``display emotional responses to prior talk in text-mediated conversations'' \cite{tolins-gifs}.
We propose a fully-automated method that captures in-the-wild texts, naturally supervised using \textit{fine-grained}, \textit{induced reaction} labels. We also augment our dataset with sentiment and emotion labels.
We use our method to collect and publish the ReactionGIF dataset\footnote{\href{https://github.com/bshmueli/ReactionGIF}{github.com/bshmueli/ReactionGIF}}.
\section{Automatic Supervision using GIFs}
\Cref{fig:2tuple} shows a typical  Twitter thread. User
$A$ writes \textit{``I can’t take this any more!''}.
User $B$ replies with a reaction GIF depicting an embrace. Our method automatically infers a  \textit{hug} reaction, signaling that $A$'s text induced a feeling of love and caring. 
In the following, we formalize our method.
\subsection{The Method}
Let $(t, g)$ represent a 2-turn online interaction with a root post comprised solely of text $t$, and a reply 
containing only reaction GIF $g$. 
Let $R=\{R_1, R_2, ..., R_M\}$ be a set of $M$ different \textit{reaction categories} representing various affective states (e.\,g., \textit{hug}, \textit{facepalm}). 
The function  $\mathfrak{R}$ maps a GIF $g$ to a reaction category, $g \mapsto \mathfrak{R}(g)$,
$\mathfrak{R}(g) \in R$.
We use $r=\mathfrak{R}(g)$ as the label of $t$.
In the Twitter thread shown in \Cref{fig:2tuple}, the label of the tweet \textit{``I can't take this any more!''} is $r = \mathfrak{R}(g) = \mathit{hug}$.

Inferring $\mathfrak{R}(g)$ would usually require humans to manually view and annotate each GIF. 
Our method automatically determines the reaction category conveyed in the GIF.
In the following, we explain how we automate  this step.

\paragraph{GIF Dictionary} We first build a dictionary of GIFs and their reaction categories by taking advantage of the 
2-step process by which users post reaction GIFs. We describe this process on Twitter; other platforms follow a similar approach:

\textit{Step 1}: The user  clicks on the \fbox{GIF} button. A menu of  reaction categories pops up (\Cref{fig:categories}). Twitter has 43  pre-defined categories (e.\,g., \textit{high five}, \textit{hug}). The user
clicks their preferred category.

\textit{Step 2}: A grid of reaction GIFs from the selected category is displayed (\Cref{fig:hugs}). The user selects one reaction GIF to insert into the tweet.

To compile the GIF dictionary, we collect the first 100  GIFs in each of the $M=43$ reaction categories on Twitter. 
We save the 4300 GIFs, along with their categories, to the dictionary. 
While in general GIFs do not necessarily
contain affective information, our method collects \textit{reaction} GIFs that depict corresponding affective states.

\paragraph{Computing $\mathfrak{R}(g)$} Given a $(t, g)$ sample, we label text $t$ with reaction category $r$ by mapping  reaction GIF $g$ back to its category $r =\mathfrak{R}(g)$. We search for $g$ in the GIF dictionary and identify the category(ies) in which it is offered to the user. If the GIF is not found, the sample is discarded. 
For the small minority of GIFs that appear in two or more categories, we look at the positions of the GIF in each of its categories and select the category with the higher position.

\begin{figure*}[t]
\centering
\includegraphics[width=0.91\linewidth]{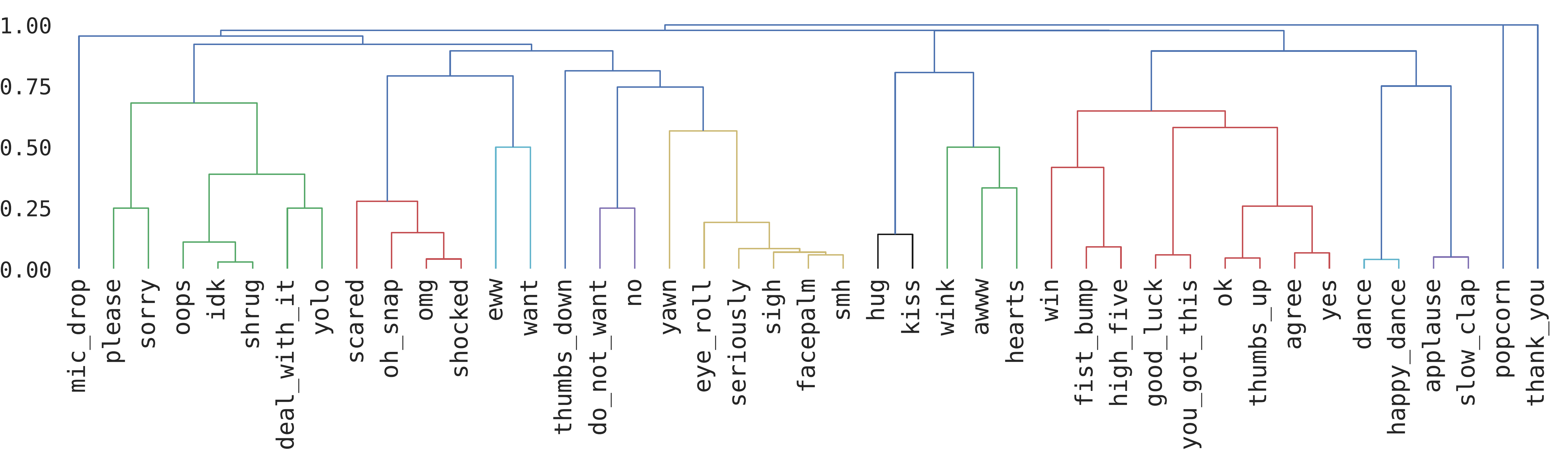}
\caption{Hierarchical clustering (average linkage) of reaction categories shows relationships between reactions.}
\label{fig:dendro}
\end{figure*}

\subsection{Category Clustering} 
\label{sec:clustering}
Because reaction categories represent overlapping affective states, a GIF may  appear in multiple categories. For example, a GIF that appears in the \textit{thumbs up} category may also appear in the \textit{ok} category, since both express approval. 
Out of the 4300 GIFs, 408 appear in two or more categories. 
Exploiting this artefact, we propose a new metric: the pairwise \textit{reaction similarity},
which is the number of reaction GIFs that appear in a pair of categories. 

To automatically discover affinities between reaction categories, we use our similarity metric and perform  hierarchical clustering with average linkage.
The resulting dendrogram, shown in \Cref{fig:dendro},
 uncovers 
 surprisingly well 
 the relationships between common human gesticulations.
 For example,  \textit{shrug} and \textit{idk} (\textbf{I} \textbf{d}on't \textbf{k}now)
 share common emotions related to uncertainty and defensiveness.
 In particular, we can see two major clusters capturing negative sentiment (left cluster: \textit{mic drop} to \textit{smh}
 [\textbf{s}hake \textbf{m}y \textbf{h}ead]) and positive sentiment (right cluster: \textit{hug} to \textit{slow clap}), which are useful for downstream sentiment analysis  tasks.
The two rightmost singletons, \textit{popcorn} and \textit{thank you}, lack sufficient similarity data.

\section{ReactionGIF Dataset}
\label{sec:dataset}
We applied our proposed method to  30K  English-language $(t, g)$ 2-turn pairs collected from Twitter in April 2020. $t$ are text-only root tweets (not containing links or media) and $g$ are pure GIF reactions. We label each tweet $t$ with its reaction category $r=\mathfrak{R}(g)$. 
See \Cref{sec:samples} for samples.
The  resulting dataset, ReactionGIF, is publicly available.

\begin{figure*}[t]
\centering
\includegraphics[width=\linewidth]{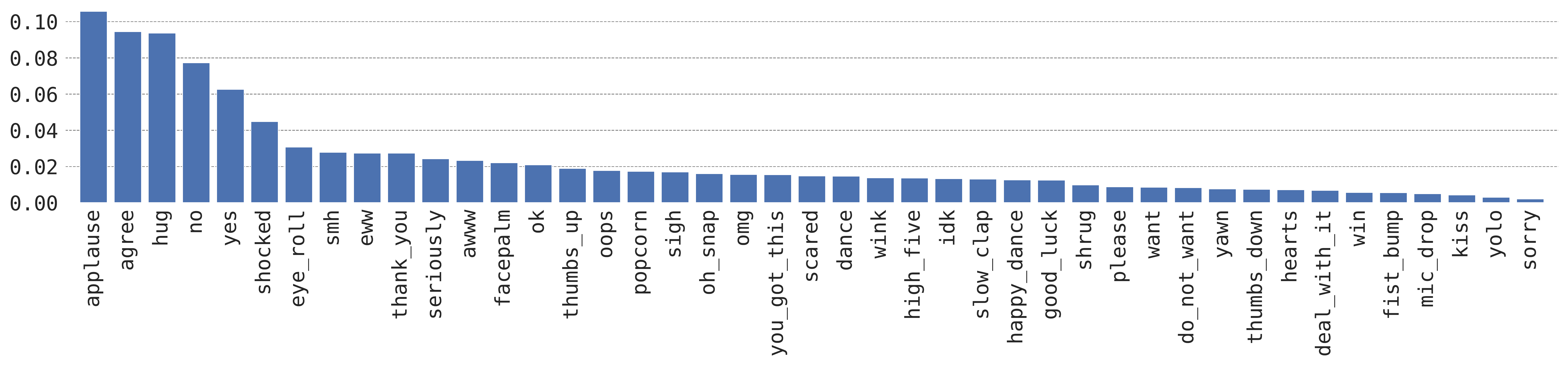}
\caption{Distribution of the 43 reaction categories in ReactionGIF}
\label{fig:distribution}
\end{figure*}

\Cref{fig:distribution} shows the category distribution's long tail. The top seven categories (\textit{applause} to \textit{eyeroll})  label more than half of the samples (50.9\%). Each of the remaining 36 categories accounts for between 0.2\% to 2.8\% of the samples.
\paragraph{Label Augmentation}
Reaction categories convey a rich affective signal. We can thus augment the dataset with other affective labels. 
We add \textbf{sentiment labels} by using the positive and negative reaction category clusters, labeling each sample according to its cluster's sentiment (\Cref{sec:clustering}). 
Furthermore, we add \textbf{emotion labels} using a novel reactions-to-emotions mapping:
we asked 3 annotators  to map
each reaction category onto a subset of the 27 emotions in
\newcite{demszky-etal-2020-goemotions}  --- see \Cref{tab:emotions}. Instructions were to view
the GIFs in each category and select the expressed emotions. Pairwise Cohen's kappa indicate moderate interrater agreements with $\kappa_{12}=0.512$, $\kappa_{13}=0.494$, $\kappa_{23}=0.449$, and 
Fleiss' kappa  $\kappa_{F}= 0.483$. We use the annotators' majority decisions as the final many-to-many mapping and label each sample according to its category's mapped emotions subset.

\begin{table}[t]
\centering
\resizebox{1.0\columnwidth}{!}{%
    \begin{tabular}{@{}llll@{}}
    \toprule
    Admiration	& Curiosity& Fear & Pride\\
    Amusement & Desire	& Gratitude & Realization\\
    Anger	&Disappointment & Grief&Relief\\ 
    Annoyance & Disapproval & Joy&Remorse\\
    Approval&  Disgust & Love&Sadness\\
    Caring & Embarrassment & Nervousness&Surprise\\ 
    Confusion & Excitement & Optimism&\\
    \bottomrule
    \end{tabular}%
}
    \caption{The 27 emotions in \citet{demszky-etal-2020-goemotions}.}
    \label{tab:emotions}
\end{table}

\paragraph{GIFs in Context} As far as we know, our dataset is the first to  offer reaction GIFs
with their eliciting texts. 
Moreover, the reaction GIFs are labeled with a reaction category. 
Other available GIF datasets (TGIF by \citealp{li2016tgif},  and 
GIFGIF/GIFGIF+, e.\,g., \citealp{jou2014predicting}) lack both the eliciting texts and the reaction categories.

\section{Baselines}
\begin{table*}[t]
\centering
\resizebox{0.61\linewidth}{!}{%
    \begin{tabular}{@{}l|cccc|cccc|cccc|c@{}}
    \toprule
     Task $\rightarrow$ & \multicolumn{4}{c|}{Reaction} &  \multicolumn{4}{c|}{Sentiment}&  \multicolumn{1}{c}{Emotion} \\ \midrule
    Model $\downarrow$& Acc & P & R & $F_1$ &                                         Acc & P & R & $F_1$&              LRAP \\ \midrule
    Majority  & 10.4 & 1.1 & 10.4 & 2.0 &                                 58.0 & 33.7 & 58.0 & 42.6 &      0.445 \\
    LR        & 22.7 & 19.5 & 22.7 & 18.0 &                                     64.7 & 64.4 & 64.7 & 62.4   &    0.529 \\
    CNN        & 25.5 & 17.3 & 25.5 & 19.1 &                                   67.1 & 66.8 & 67.1 & 66.3
   &    0.557 \\
    RoBERTa      & \textbf{28.4} & \textbf{23.6} & \textbf{28.4} & \textbf{23.9} &                                               \textbf{70.0} & \textbf{69.7} & \textbf{70.0} & \textbf{69.8} &                   \textbf{0.596}       \\ \bottomrule
    \end{tabular}%
}
    \caption{Baselines for the reaction, sentiment, and emotion classification tasks. 
    All metrics are weight-averaged.   
    The highest value in each column is emboldened.}
    \label{tab:experiments}
\end{table*}

As this is the first dataset of its kind, we aim to promote future research by offering baselines for predicting the reaction, sentiment, and emotion induced by tweets. We use the following  four models in our experiments: 

\begin{itemize}
    \item \textbf{Majority}: A simple majority class classifier.
    \item \textbf{LR}: Logistic regression classifier (L-BFGS solver with $C=3$, maximum iterations 1000, 
    stratified K-fold cross validation with $K=5$) using TF-IDF vectors (unigrams and bigrams, cutoff 2, maximum 1000 features, removing English-language stop words).
    \item \textbf{CNN}: Convolutional neural network (100 filters, kernel size 3, global max pooling; 2 hidden layers with 0.2 dropout; Adam solver, 100 epochs, batch size 128, 
    learning rate 0.0005) with GloVe embeddings (Twitter, 27B tokens, 1.2M vocabulary, uncased, 100d)
    \cite{pennington2014glove}.
    \item \textbf{RoBERTa}: Pre-trained transformer model (base, batch size 32, maximum sequence length 96, 3 training epochs)
    \cite{liu2019roberta}. 
\end{itemize}

We hold out 10\% of the samples for evaluation. 
The code is publicly available along with the dataset for reproducibility.
The experiment results are summarized in \Cref{tab:experiments}.

\textbf{Affective Reaction Prediction} is a multiclass classification task where we predict the reaction category $r$ for each tweet $t$. RoBERTa achieves a weight-averaged $F_1$-score of 23.9\%. 

\textbf{Induced Sentiment Prediction} is a binary classification task to predict the sentiment induced by tweet $t$ by using the augmented labels. RoBERTa has the best performance
 with accuracy 70.0\% and $F_1$-score of 69.8\%. 
  
Finally, \textbf{Induced Emotion Prediction} uses our reaction-to-emotion transformation for predicting
emotions. This is a 27-emotion \textit{multilabel} classification task, reflecting our dataset's
unique ability to capture complex emotional states. RoBERTa is again the best model, with Label Ranking Average Precision (LRAP) of 0.596.

\section{Discussion}
Reaction GIFs are ubiquitous in online conversations due to their uniqueness as lightweight
and silent moving pictures. They are also more effective and precise\footnote{For example, the \textit{facepalm} reaction is ``a gesture in which the palm of one's hand is brought to one's face, as an expression of disbelief, shame, or exasperation.'', Oxford University Press, \href{https://lexico.com/en/definition/facepalm}{lexico.com/en/definition/facepalm}} 
when conveying affective states compared to text, emoticons, and emojis \cite{bakhshi2016fast}.
Consequently, the reaction category is a new type of label,
not yet available in NLP emotion datasets:
existing datasets 
use either the  discrete emotions model \cite{ekman1992argument} or the dimensional  model of emotion
\cite{mehrabian1996pleasure}. 
The new labels
possess important advantages, but also present interesting challenges. 

\paragraph{Advantages} 
The new reaction labels provide a rich, complex signal that can be mapped to other types of affective labels, including sentiment,  emotions and possibly feelings and moods. In addition, because reaction GIFs are ubiquitous in online conversations, we can automatically collect large amounts of
inexpensive, naturally-occurring, high-quality affective labels. Significantly, and in contrast with most other emotion datasets, the labels 
measure \textit{induced} (as opposed to \textit{perceived}) affective states; these labels are of prime
importance yet the most difficult to obtain, with applications that include GIF recommender systems, dialogue systems, and any other
application that requires predicting or inducing users' emotional response.

\paragraph{Challenges}
The large number of reaction categories (reflecting
the richness of communication by gestures) makes their prediction a challenging task. In addition, the category distribution has a long tail, and there is an affective overlap between the categories. 
One way to address these issues is by accurately mapping the reactions to emotions.
Precise mapping  will require a larger GIF dictionary (our current one has 4300 GIFs), a larger
dataset, and new evaluation metrics. A larger GIF dictionary will also improve the \textit{reaction similarity}'s accuracy, offering new approaches for studying relationships between reactions (\Cref{sec:clustering}). 

\section{Conclusion}
Our new method is the first to exploit the use of reaction GIFs for capturing in-the-wild \textit{induced} affective data. We augment the data with induced sentiment and emotion labels using two novel mapping techniques: reaction category clustering and reactions-to-emotions transformation. 
We used our method to publish ReactionGIF, a first-of-its-kind dataset with
multiple affective labels. The new method and dataset offer opportunities for advances in emotion detection.

Moreover, our method can be generalized to capture data from other social media and instant messaging platforms that use reaction GIFs, as well as applied to other downstream tasks
such as multi-modal emotion detection and emotion recognition in dialogues, thus  enabling new research directions in affective computing.
\section*{Acknowledgements}
We thank the anonymous reviewers for their
comments and suggestions.
Special thanks to Thilina Rajapakse, creator of the elegant Simple Transformers package, for his help. 

This research was partially supported by the Ministry of Science and Technology in Taiwan under grants MOST 108-2221-E-001-012-MY3 and MOST 109-2221-E-001-015- [\textit{sic}].

\section*{Ethical Considerations and Implications}
\subsection*{Data Collection}
The ReactionGIF data was collected from Twitter using the official API in full accordance with their Development Agreement and Policy \cite{Develope14:online}.  Similar to other Twitter datasets, we include the tweet IDs but not the texts. This guarantees that researchers who want to use the data will also need to agree with Twitter's Terms of Service. It also ensures compliance with section III (Updates and Removals) of the Developer Agreement and Policy's requirement that when users delete tweets (or make them  private), these changes are reflected in the dataset \cite{belli2020privacy}.

\subsection*{Annotation}
Annotation work was performed by three adult students, two males and one female, who use social media regularly. The labeling involved viewing 43 sets of standard reaction GIFs, one for each reaction category. These reaction GIFs are the standard offering by the Twitter platform to all its users. As a result, this content is highly familiar to users of social media platforms such as Facebook or Twitter, and thus presents a very low risk of psychological harm. Annotators gave informed consent after being presented with details about the purpose of the study, the procedure, risks, benefits, statement of confidentiality and other standard consent items. Each annotator was paid US\$18. The average completion time was 45 minutes.

\subsection*{Applications}
The dataset and resulting models can be used to infer readers' induced emotions. Such capability can be used to help online platforms detect and filter out content that can be emotionally harmful, or emphasize and highlight texts that induce positive emotions with the potential to improve users' well-being. For example, when a person is in grief or distress, platforms can give preference to responses which will induce a feeling of caring, gratitude, love, or optimism. Moreover, such technology can be of beneficial use in assistive computing applications. For example, people with emotional disabilities can find it difficult to understand the emotional affect in stories or other narratives, or decipher emotional responses by third parties. By computing the emotional properties of texts, such applications can provide hints or instructions and provide for smoother and richer communication. However, this technology also has substantial risks and peril. Inducing users' affective response can also be used by digital platforms in order to stir users into specific action or thoughts, from product purchase and ad clicking to propaganda and opinion forming. Deployers must ensure that users understand and agree to the use of such systems, and consider if the benefit created by such systems outweigh the potential harm that users may incur.

\bibliographystyle{acl_natbib}
\bibliography{reaction}
\begin{figure*}[t!]
\centering
\includegraphics[width=0.91\linewidth]{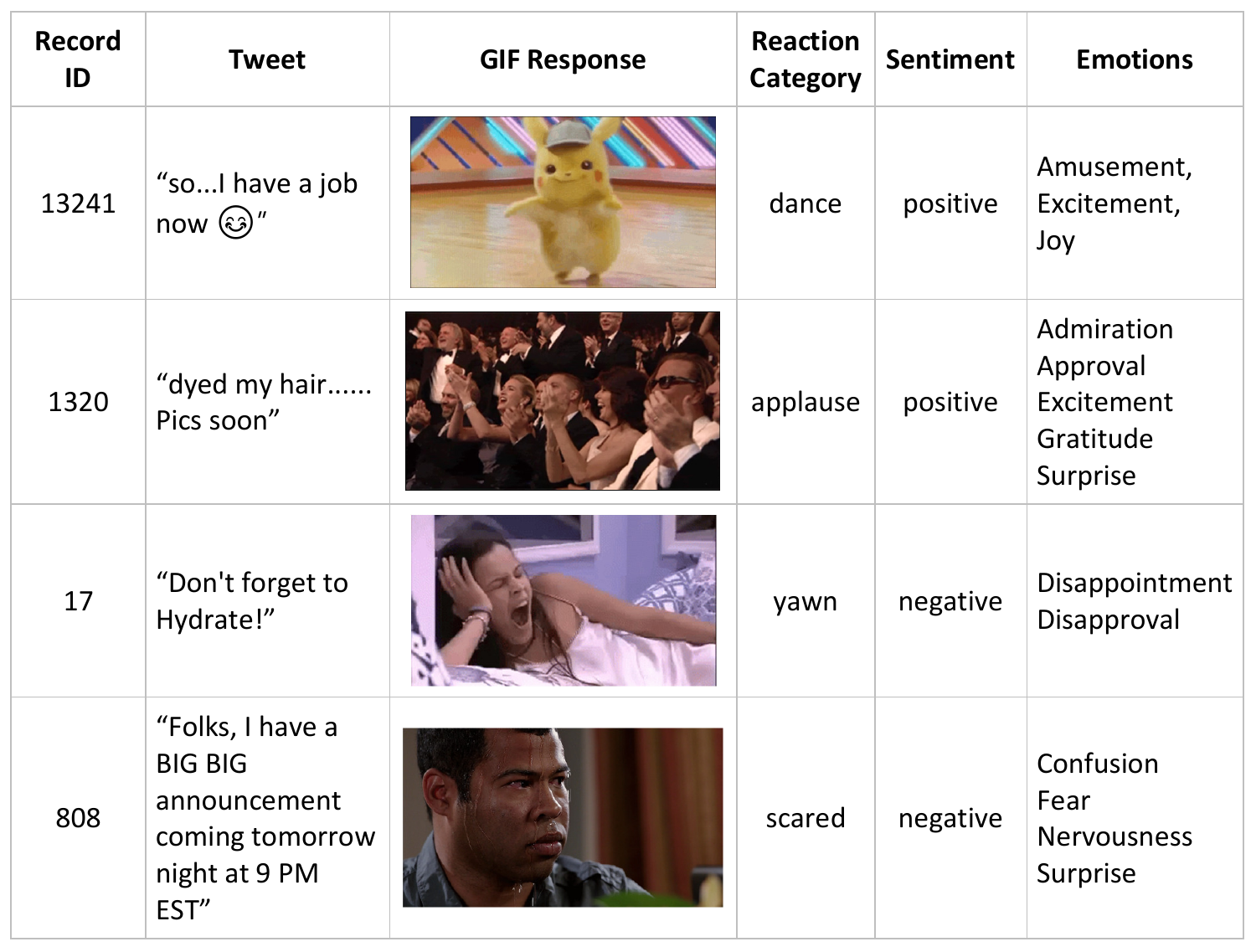}
\caption{ReactionGIF samples.}
\label{fig:samples}
\end{figure*}

\appendix
\section{Dataset Samples}
\label{sec:samples}
\Cref{fig:samples} includes four samples from the dataset. For each sample, we show the record ID within the dataset, the text of the tweet, a thumbnail of the reaction GIF, the reaction category of the GIF, and the two augmented labels: the sentiment and the emotions.
\end{document}